\documentclass[12pt]{article}

\usepackage{scicite}

\usepackage[rightcaption]{sidecap}
\sidecaptionvpos{figure}{c}
\usepackage{epsfig}
\usepackage{graphicx}
\usepackage{amsmath}
\usepackage{amssymb}
\usepackage{wrapfig}
\usepackage{lipsum}
\usepackage{caption}
\usepackage{subcaption}

\usepackage{times}
\usepackage{physics}
\usepackage{amsfonts}
\usepackage{xcolor}
\usepackage{makecell}
\usepackage{textcomp}
\usepackage{gensymb}
\usepackage{hyperref}

\definecolor{azure(colorwheel)}{rgb}{0.0, 0.5, 1.0}
\hypersetup{
    colorlinks=true,
    urlcolor=azure(colorwheel),
}

\usepackage{environ}

\newenvironment{sciabstract}{%
\begin{quote} \bf}
{\end{quote}}

\newcounter{lastnote}

\title{Full-Body Visual Self-Modeling of \\ Robot Morphologies}

\author
{Boyuan Chen$^{\ast}$, Robert Kwiatkowski, Carl Vondrick, Hod Lipson\\
\\
\href{https://robot-morphology.cs.columbia.edu/}{robot-morphology.cs.columbia.edu}\\
Columbia University\\
\\
\normalsize{$^\ast$To whom correspondence should be addressed; E-mail:  bchen@cs.columbia.edu.}
}

\date{}

\begin{document} 


\baselineskip18pt


\maketitle 

\begin{sciabstract}

Internal computational models of physical bodies are fundamental to the ability of robots and animals alike to plan and control their actions. These ``self-models'' allow robots to consider outcomes of multiple possible future actions, without trying them out in physical reality. Recent progress in fully data-driven self-modeling has enabled machines to learn their own forward kinematics directly from task-agnostic interaction data. However, forward-kinema\-tics models can only predict limited aspects of the morphology, such as the position of end effectors or velocity of joints and masses. A key challenge is to model the entire morphology and kinematics, without prior knowledge of what aspects of the morphology will be relevant to future tasks. Here, we propose that instead of directly modeling forward-kinematics, a more useful form of self-modeling is one that could answer space occupancy queries, conditioned on the robot's state. Such query-driven self models are continuous in the spatial domain, memory efficient, fully differentiable and kinematic aware. In physical experiments, we demonstrate how a visual self-model is accurate to about one percent of the workspace, enabling the robot to perform various motion planning and control tasks. Visual self-modeling can also allow the robot to detect, localize and recover from real-world damage, leading to improved machine resiliency. 

\end{sciabstract}

\clearpage

\subsection*{Main Text}

Building computational self-models of robot bodies, or the ability of a robot to simulate its physical self, is an essential requirement for robot motion planning and control. Similar to humans and animals \cite{gallup1982self, rochat2003five}, robots can use self-models to anticipate future outcomes of various motion plans without explicitly trying them out in the physical world. The predictions via self-models can be utilized in decision criteria of future actions. Importantly, a consistent self-model, once acquired, can be re-purposed to many different of tasks, and thus can serve for lifelong learning.

Most available robotic systems rely on dedicated physical simulators \cite{koenig2004design, todorov2012mujoco, faure2012sofa, coumans2016pybullet, lee2018dart, drake} for task planning and control. Yet, these simulators require extensive human effort to develop. In contrast, recent progress in fully data-driven self-modeling has enabled machines to learn their forward kinematics directly using task-agnostic interaction data. 

Data driven forward-kinematics self-models typically need to know in advance what aspects of the robot need to be modeled, such as the tilt angle of the robot \cite{bongard2006resilient}, the position of end effectors \cite{kwiatkowski2019task}, the velocity of motor joints \cite{sanchez2018graph}, the mirror image of animatronic faces \cite{9560797}, or the contact locations as well as joint configurations of robot grippers \cite{hang2021manipulation}. The restricted predictive scope of traditional data-driven self-models limits the general applicability of these self-models to future, yet unknown, 3D spatial planning tasks. For example, a data-driven self model focusing only on predicting the position of an end effector, may not be useful for tasks involving operation in a crowded workspace, where full body collisions must be factored into the planning. Making sure that the entire robot arm motion will be collision free is a critical aspect for numerous safe robot operations such as object retrieval, trajectory planning and human-robot interaction. Modeling the entire robot morphology and kinematics, without prior knowledge of what aspects of the morphology are relevant to future tasks, has remained a major challenge.

\begin{figure*}[t!]
    \centering
    \includegraphics[width=0.95\linewidth]{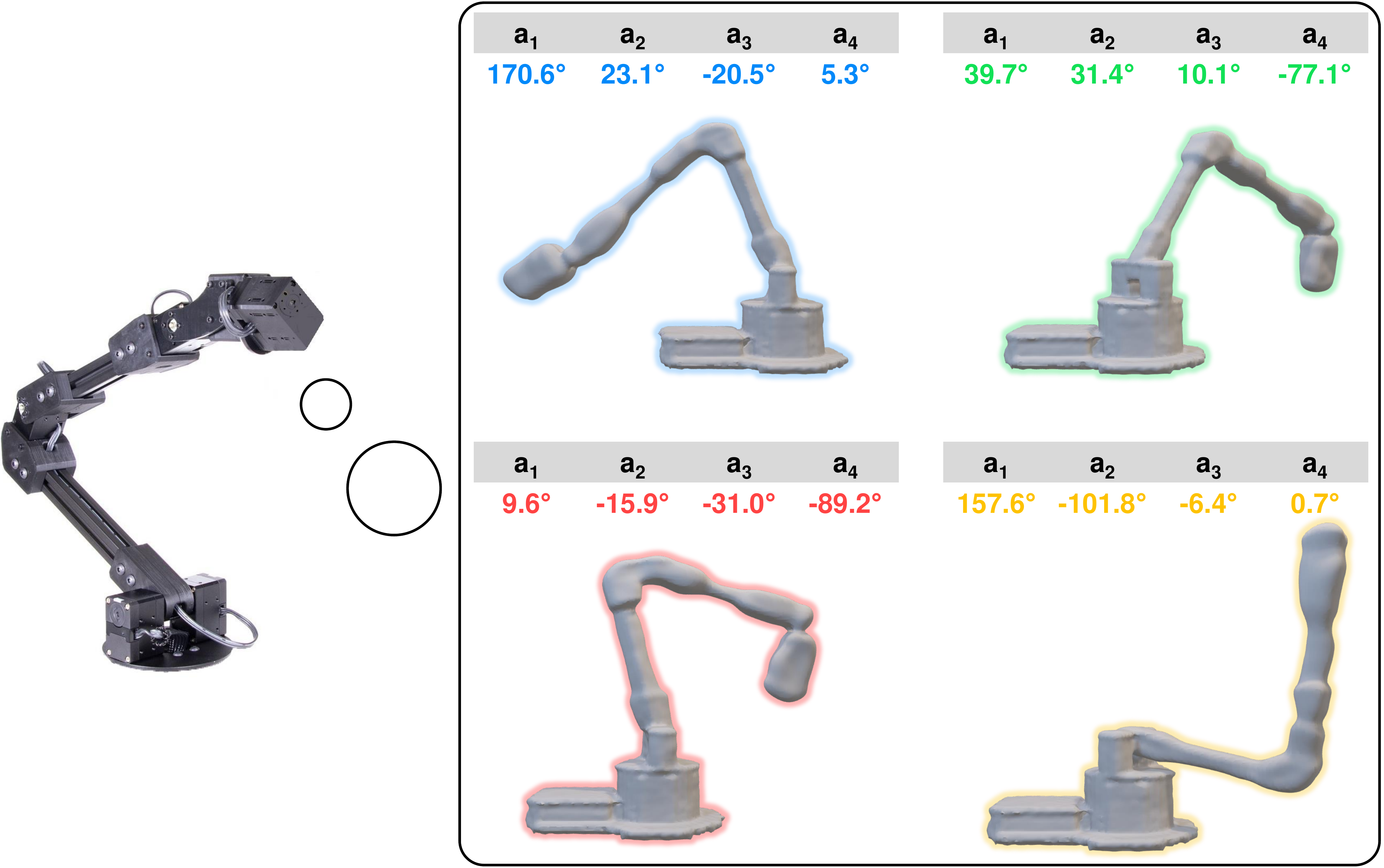}
    \caption{\textbf{Visual self-modeling robots.} We equip the robot with the ability to model its entire morphology and kinematics in 3D space only given joint angles, known as visual self-model. With the visual self-model, the robot can perform variety of motion planning and control tasks by simulating the potential interactions between itself and the 3D world. Our visual self-model is continuous, memory efficient, differentiable and kinematic aware.}
    \label{fig:teaser}
\end{figure*}

In this paper, we present a full-body visual self-modeling approach (Fig.\ref{fig:teaser}) which captures the entire robot morphology and kinematics using a single implicit neural representation. This implicit system is able to answer space occupancy queries, given the current state (pose) or the possible future states of the robot. For example, the query-driven visual self-model can answer queries as to whether a spatial position $(x, y, z)$ will be occupied if the joints move to some specified angles. Since both the spatial and robot state inputs are real values, our visual self-model allow continuous queries in the domain of both control signals and spatial locations. Furthermore, the learning process only requires joint angles and sparse multi-view depth images, which enables generalizable and scalable data acquisition without human supervision.

Once learned, the responses from this single visual self-model to a series of queries can then be used for a variety of 3D motion planning and control tasks, even though the self-model was only trained with task-agnostic random motor movements. Because of our fully differentiable parametrization, the robot can directly perform efficient parallel gradient-based optimization on top of the self-model to search for the best plans in real time. We can also combine the self-model in a plug-and-play manner with existing motion planners. Moreover, when the robot sustains physical damage, such as broken motor or changed topology, our self-model can detect, identify and recover from these changes. Since our self-model is inherently visual, it can provide a real-time human-interpretable visualization of the robot's internal belief of its current 3D morphology. This ability to sense pose-conditioned space occupancy is similar to our natural human ability to ``see in our minds eye'' \cite{gardner2011frames} whether our body could fit through a narrow passage, without actually trying it out.

In the following sections, we begin by describing our main methods for constructing the visual self-model. We follow up with describing ways in which the robot can re-use the learned self-model in multiple downstream tasks. We then discuss the pipelines of damage detection, recognition and recovery.

\subsubsection*{Implicit visual self-model Representation}

Robots operate in a 3D world, and therefore being morphologically and kinematically aware in 3D space is essential for them to successfully interact with the physical environments as well as adapt to potential changes in the field. Traditionally, robot engineers build a physical simulator and integrate it with CAD models of the robot. However, designing a simulation environment is not trivial. Accurate CAD models that reflect the real as-built robot geometry may not be easily available, especially for robots that have been modified due to damage, adaptation, wear and repair. This challenge will likely become more acute as the variety and complexity of robotic systems continues to increase in the future, and especially as robots must operate with less human supervision, maintenance, and oversight.

We therefore aim to learn the self-model of robots directly through task-agnostic data with minimal human supervision or domain knowledge. Our goal is to learn a visual self-model which can capture the entire body morphology and kinematics, without prior knowledge of the body configurations such as joint placements, part geometry, motor axis and joint types. With the visual self-model, a robot should be able to plan its future actions by rolling out the self-model before executing any actions in the physical world. We can also visualize its final plan from different viewing angles, because the model itself is three-dimensional.

\begin{figure*}[t!]
    \centering
    \includegraphics[width=\linewidth]{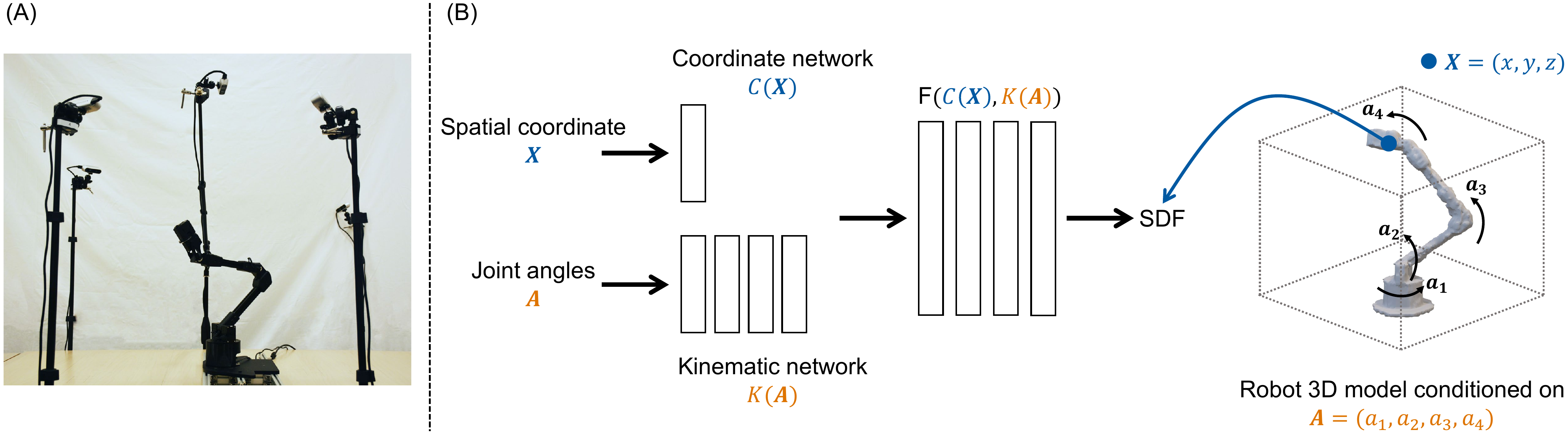}
    \caption{\textbf{Implicit visual self-model representation.} (A) Real-world setup for data collection. We fused sparse views from five depth cameras to capture the point cloud of the robot body. As the robot arm randomly moved around, we recorded pairs of the robot joint angles and its 3D point cloud. See S1 Movie for real-time data collection. (B) We show the computational diagram of our visual self-model. The coordinate network takes in the spatial coordinate and the kinematic network extracts kinematic features from the input joint angles. We then concatenated the spatial features and the kinematic features into a few layers of MLPs to output the zero-level set SDF values. The implicit representation can be queried at arbitrary continuous 3D spatial coordinates and different sets of joint angles.}
    \label{fig:diagram}
\end{figure*}

There are two major challenges when designing a visual self-modeling process. First, we need to carefully decide how to represent the 3D geometry of the robot body. Most existing 3D representations are explicit, such as point cloud, tessellated triangle meshes, or voxelized occupancy grids. However, such approaches come with several limitations. Point clouds, meshes, and grids often consume large amounts of memory to store even a single geometry, let alone a kinetic geometry dependent on input DoF. Point clouds also lose structural connectivity, while voxel representations lose continuous resolutions. These limitations are amplified in kinematic tasks, since the visual self-models are expected to be dependent on trajectories of multiple degrees of freedom of the robot.

The second challenge concerns the computational efficiency of leveraging the learned visual self-model for downstream task planning. Once a visual self-model is formed, we hope that the same model can be used for many tasks. In other words, the model must be task-agnostic. Furthermore, real-time planning and control is critical for many robotic applications. Therefore, the ideal representation should render the 3D model in a parallel and memory efficient manner using GPU hardware. The model should also provide fast inference capability to solve common inverse problems in robotics, such as inverse kinematics. Lastly, not every component of the robot body weights equally in all tasks, so it should be possible to query different spatial components of the visual self-model as needed. For example, the full 3D knowledge of the robot base and 3D geometry of other arm components are not required when calculating the inverse kinematic solution of a robot arm trying to reach a 3D object with its end effector.

We overcame the above challenges by proposing a state conditioned implicit visual self-model that is continuous, memory efficient, differentiable, and kinematics aware. The key idea is that the model does not simply predict future robot states explicitly; instead, it is able to answer spatial and kinematic queries about the geometry of the robot under various future states.

To construct a query-answering self-model, we leverage implicit neural representations to model the 3D body of the robot as shown in Fig.\ref{fig:diagram}. Given a spatial query point coordinate $\vb*{X} \in \mathbb{R}^3$ normalized based on scene boundary, and a robot joint state vector $\vb*{A} \in \mathbb{R}^N$ specifying all the $N$ joint angles, the visual self-model can be represented by a neural network to produce the zero-level set Signed Distance Function (SDF) of the robot body at the given query point $\vb*{X}$. Formally, the model can be expressed as:

\[ SDF = \{\vb*{X} \in \mathbb{R}^3, \vb*{A} \in \mathbb{R}^N | F(C(\vb*{X}), K(\vb*{A}))\}, \]

where $C$ is the coordinate neural network with several layers of MLPs to encode the spatial coordinate features, $K$ is the kinematic neural network with several layers of MLPs to encode the robot kinematic features, and $F$ is the last few layers of MLPs to fuse the features from both the coordinate network $C$ and kinematic network $K$ after concatenating their outputs to produce the final SDF values conditioned on the queried spatial coordinates and current joint angles. We omit the batch size here for simplicity. For nonlinear activation functions, we used Sine functions to preserve the details on the 3D models \cite{sitzmann2020implicit}.

We trained the network by formulating the problem as an Eikonal boundary value problem. Instead of supervising the network with ground-truth SDF, similar to SIREN network \cite{sitzmann2020implicit}, we directly used point clouds and surface normals obtained by fusing observations from sparse RGB-D camera views as labels as indicated in Fig.\ref{fig:diagram}. In both simulation and real-world setup, we used five RGB-D cameras to capture pairs of data for training, namely the joint angles and the fused point cloud. During testing, the only available robot-related information to our visual self-model is a set of joint angles. The detailed loss function is discussed in the Methods section.

Overall, our visual self-model is formed by several layers of MLPs that implicitly captures the entire morphology and kinematics of the robot body. We implemented the network with differentiable deep learning framework so that it can be easily deployed on GPUs with end-to-end differentiable capabilities. Notably, though the entire self-model only consumes 1.1 MB memories to store its weights, our visual self-model can represent the 3D morphology of the robot body with different kinds of joint angles at various continuous spatial locations. By separating the kinematic feature encoder and coordinate feature encoder into two sub-networks, each sub-network captures independent semantic meaning. As we will show next, this property allows the self-model to learn rich kinematic features useful for downstream tasks.

\begin{figure*}[t!]
    \centering
    \includegraphics[width=\linewidth]{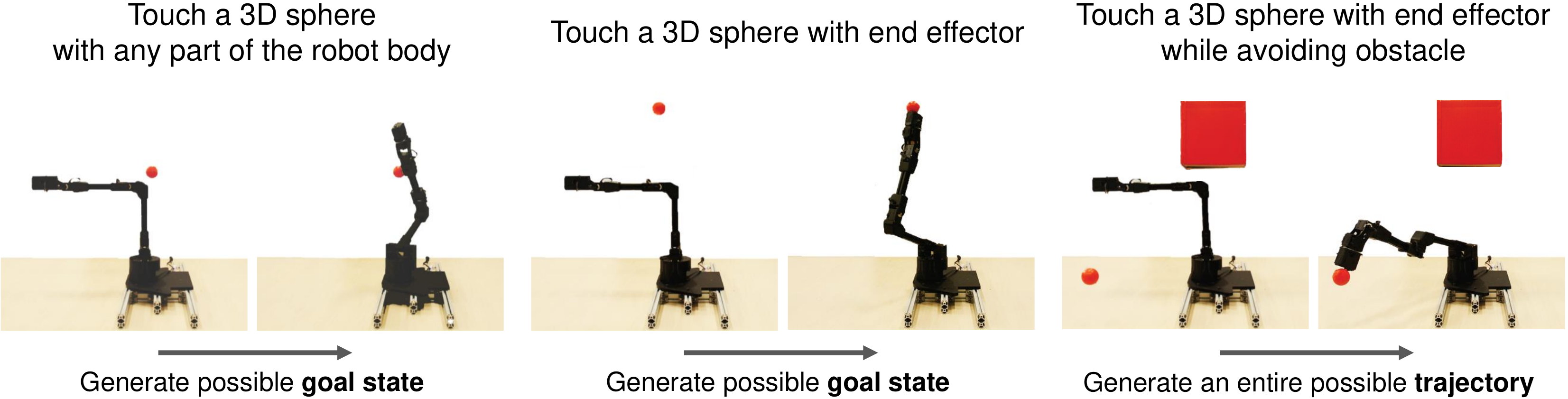}
    \caption{\textbf{3D self-aware motion planning tasks.} We present an overview of three different tasks. Touch 3D sphere with any part of the robot body (Left) asks the robot to generate a set of target joint angles such that some part of the robot body needs to be in contact with a randomly placed target sphere. Touch a 3D sphere with end effector (Middle) requires the robot to generate a set of target joint angles such that the robot needs to touch a randomly placed target sphere with its end effector link. Touch a 3D sphere with end effector while avoiding obstacle (Right) tasks the robot to propose an entire set of collision-free trajectories in the form of intermediate joint angles to touch a randomly placed target sphere using its end effector. The three tasks gradually becomes harder with more constraints.}
    \label{fig:task-overview}
\end{figure*}

\subsubsection*{3D Self-Aware Motion Planning}

We aim to utilize the learned visual self-model in various motion planning tasks in 3D space. In this section, we will present algorithm designs to show the use cases for three sample tasks (Fig. \ref{fig:task-overview}). However, our model is not limited by only those three tasks. Rather, we use them as representative examples for demonstration purposes and we expect that the model can generalize to other possible tasks.

\begin{itemize}
    \item \textbf{Touch a 3D sphere with any part of the robot body} The goal of this task is to have the robot to touch a 4cm diameter sphere using any part of its body. To solve this problem, the robot needs to calculate inverse kinematics in the 3D space without constraints on which specific body piece touches the target object.
    \item \textbf{Touch a 3D sphere with end effector} This task not only requires the robot to touch a sphere with about diameter of 4cm, but it also asks the robot to touch the target object with its end effector link. This is a harder task since the robot needs to solve inverse kinematics in the 3D space with a particular link constraint. The solution space is quickly reduced.
    \item \textbf{Touch a 3D sphere with end effector, while Avoiding an Obstacle} In this task, we ask the robot to go beyond computing a target end state with or without link constraints. Instead, in order to succeed at this task, the robot needs to perform precise motion planning in 3D to touch the final target while avoiding a large obstacle shown as red block in Fig.\ref{fig:task-overview}. Overall, the robot is tasked to propose an entire safe trajectory from its initial state to the target state. During the execution of the proposed trajectory, the robot will fail the task if any part of the robot body collides with the obstacle.
\end{itemize}

To solve the these motion planning tasks, one immediate thought is to obtain the entire robot body meshes and load them into existing robot simulators. This can be done by traversing all possible spatial points under certain precision and different sets of joint angles through the implicit neural representation, and rendering the entire 3D mesh of the robot body with post-processing algorithms \cite{lorensen1987marching}. This usage of the visual self-model seems to be a straightforward solution to bypasses the need to construct robot kinematic and geometric models such as CAD and URDF files. However, in practice, we found that constantly loading new robot meshes and destroying old robot meshes in commonly available robot simulators costed a significant amount of time. This limits the possibility of applying this method for real-time planning and control.

We propose to frame the first two tasks as constrained optimization problems by leveraging the differentiability of the visual self-model as well as its capability of answering partial queries on spatial coordinates. Specifically, for the first task, we initialize thousands sets of joint angles. We then sample $P$ points uniformly on the surface of the target object.  Since the output of the visual self-model will be zero when the queried spatial point is on the surface, the overall objective is to find the set of joint angles which can minimize the total sum of output values across all the sampled points on the target object. By freezing the weights of the learned visual self-model, we can perform gradient descent from the output surface predictions with respect to the input joint angles, under the constraint that the motor angles has to be within the range of $[-\pi, \pi]$.

Formally, the constrained optimization problem can be expressed as:
\[ \vb*{A}^* =  \min  \mathbb{E}_{\vb*{A}^b} \left[ \Sigma_{\vb*{T}^p} F(C(\vb*{T}^p), K(\vb{A}^b)) \right] \quad \textrm{\textbf{s.t.}} -\pi \leqslant A_{i}^{b} \leqslant \pi , i=1, 2, 3, 4 \]

where $\vb*{T} \in \mathbb{R}^{P \times 3}$ is the sampled points on the target object, $i$ is the motor index, and $b=1, 2, ..., B$ is the index of each sampled set of joint angles with the maximum value $B$ to be the batch size on a single GPU. Since the visual self-model runs parallelly on a GPU with small consumption of memories, the entire optimization process can produce accurate solutions within a short period of time. With more GPUs, the process can be further sped up.

To solve the second task, we need additional information about where the end effector locates relative to the entire robot body. Since the current visual self-model was only trained to capture the overall body geometries, similar to other works in self modeling, we can supervise the visual self-model to predict the end effector location at the same time. The good news is that our visual self-model already has a specialized sub-network that implicitly captures the robot kinematics. Therefore, we can directly use the pretrained weights of the kinematic sub-network, and train only two nonlinear layers of MLPs $E$ attached to the end of the sub-network with little additional efforts. In fact, as we will show in the experiment section, our visual self-model provides a strong semantic proxy to pre-train the kinematic sub-network, leading to superior performance than training a specialized network to predict the end effector position from scratch. Without our decomposition formulation of kinematic sub-network, the acquired kinematics information may not be easily distilled as an independent feature for future use.

Similar to the first task, we now can formalize the solution of the second task by adding another objective function to make sure the resulted end effector reaches the target object. The overall optimization problem can be formalized as follows:

\[ \vb*{A}^* =  \min  \mathbb{E}_{\vb*{A}^b} \left[ \Sigma_{\vb*{T}^p} \lambda_{\textrm{EE}} E(K(\vb{A}^b)) + \lambda_{\textrm{SDF}} F(C(\vb*{T}^p), K(\vb{A}^b)) \right] \textrm{\textbf{s.t.}} -\pi \leqslant A_{i}^{b} \leqslant \pi , i=1, 2, 3, 4 \]

As discussed above, the objective function includes two terms weighted by hyperparameters $\lambda_{\textrm{EE}}$ and $\lambda_{\textrm{SDF}}$. The first term ensures the end effector touches the target object and the second term encourages the robot body to touch the target object. We found that adding a small $\lambda_{\textrm{SDF}}$ consistently achieves better results.

Regarding the third task, our visual self-model can directly work with existing motion planning algorithms with minimal changes. There have been great success \cite{lavalle2006planning} on motion planning algorithms to solve obstacle avoidance problem in high-dimensional state and action spaces. We thus combine our visual self-model with the existing algorithms in a plug-and-play manner. Specifically, we use RRT* \cite{karaman2010incremental} as our backbone algorithm due to its popularity, probabilistic completeness and computational efficiency. Generally speaking, there are two major components in RRT* that require physical inference with robot bodies. The first component is to calculate the goal state, and the second component is to check whether a collision will happen given a particular state of the robot. With these two components, various planning algorithms can narrow the search space to the final solution without having to explicitly query robot status again.

Traditionally, these two components require a dedicated robot simulator and pre-defined robot bodies. With our visual self-model, we can reach the final solution by simply performing fast parallel inference on the learned model. Specifically, the goal state can be obtained by running the same optimization procedure as in the second task. For collision detection, we can pass uniformly sampled points on the obstacle surface as well as the given set of joint angles $A_{query}$ through our visual self-model as shown below. If the total sum of the output values over all the sampled points is equal to or below a small threshold $\tau$ , then there is a collision. Otherwise, the robot will not collide with the obstacle object.

\[ \textrm{Collision}=\begin{cases} \textrm{True}, \ \Sigma_{\vb*{O}^i} F(C(\vb*{O}^i), K(\vb*{A}_{query})) \leqslant \tau;\\ \textrm{False}, \ \Sigma_{\vb*{O}^i} F(C(\vb*{O}^i), K(\vb*{A}_{query})) > \tau. \end{cases}\]

\subsubsection*{Damage Identification and Recovery}

One major promise of machines that can model or identify themselves is the capability of recognizing and inspecting damage or changes, and then quickly adapting to these changes. In this section, we present our method to identify and recover from damage using the learned visual self-model.

Our approach involves three steps. The robot first detects a damage or change on its body compared to its original (intact) geometry. Then the robot can identify which specific type of damage or change is happening. Finally, the robot will gather new information about itself with limited data and computational resources to quickly adapt its self-model to the new changes.

Overall, our approach introduces several significant advantages over previous methods. First, being able to recognize the specific type of damage or change enables the robot to provide additional feedback information. Previous works have shown that it was possible to detect a damage. However, they were not able to provide additional information to identify the source of the change or which specific type of damage has happened. This information is extremely helpful when the damage requires hardware repair. Instead of relying on a domain expert to perform a series of inspections, our method can automatically generate information about specific damage such as ``the second joint motor is broken''.

Another advantage is that our approach performs modeling in the 3D visual world. This means that we can visualize and render the internal belief of the visual self-model in a straightforward and interactive fashion. As we will show in the results, one can immediately tell which section of the internal belief of the robot body does not match the real-world counterpart. We can further tell visually if the internal belief has been updated to match the new changes after learning from new observations.

In the following sections, we begin by describing the specific algorithms, and then we follow with real-world results in the next section. In the first step, we measure the current prediction error and the original prediction error. The current prediction error is computed by comparing the internal belief expressed by the learned visual self-model with the current observed 3D mesh of the robot, while the original prediction error is computed by comparing the same internal belief with the previously observed 3D robot body. Both cases share the same joint conditions. By comparing these two prediction errors, a large gap can inform us about a significant change or damage to the robot body.

In the second step, we aim to identify the specific type of damage happening on the robot. Based on the robot arm platform we are experimenting with, we assume two types of potential changes: (1) broken motor and (2) changed topology.

To reveal which specific type of the current damage is, our key idea is to solve the inverse problem with the learned visual self-model. Concretely, based on a single current observation of the robot body, we infer the best joint angles that the robot should have executed to result to the current 3D observation. This is a very challenging problem because an ideal joint angle set needs to give accurate 3D reconstruction of the entire robot body. Relying on previous gradient-based optimization algorithm is inefficient since the final gradient computation requires the sum over all the sampled points on the whole robot body. This process takes a large amount of memory and computation resources to perform a single gradient step due to the large volume of the robot mesh. Instead, we propose to use random search to locate the best possible joint angles. The simple random search algorithm works very well in this case. It does not require the accumulation of any gradient information so that larger batch of queries can fit on a single forward pass of visual self-model.

With the inferred joint angles, we can quantify the damage by comparing them with the actual input motor commands. If a specific inferred joint angle is always different from the actual input and that particular inferred angle always stays as a constant value or some other random values, then we can tell that the corresponding motor is broken. When all the inferred joint angles match closely to the actual input commands, the wrong belief of the 3D body then comes from a topology change and all the motors function well. We leave the research where both changes happen at the same time or more complex changes as future directions.

Finally, we also evaluate if our visual self-model can quickly recover from the changes by adapting on several new observations. For this step, the main purpose is to demonstrate the resiliency of the model, rather than proposing a new algorithm for continual adaption. Therefore, we follow common approaches by collecting a few more 3D observations after the changes to keep training the network for several epochs. We then check if the new visual self-model can successfully update its internal belief to match the current robot body both quantitatively and qualitatively.

\subsection*{Results}

In this section, we aim to evaluate the performance of the learned visual self-model, demonstrate the results of using the visual self-model in various motion planning tasks, and test the resiliency of the learned model under real-world damages. To this end, in the first two subsections, we present quantitative and qualitative evaluations as well as baseline comparisons in simulation. For all three subsections, we also demonstrate the fidelity of directly learning and using the visual self-model in the real-world setup.

\subsubsection*{Visual self-model Estimation}

We used the WidowX 200 Robot Arm as our experimental platform both in simulation and real-world. In order to obtain the ground truth point cloud data, we mounted five RealSense D435i RGB-D cameras around the robot as shown in Fig.\ref{fig:diagram}(A). Four cameras were around each side of the robot to capture side views. One camera was on the top to capture the top-down view. All cameras were calibrated. The depth images were first projected to point clouds which were then fused into a single point cloud based on the camera extrinsic parameters. The final point cloud was generated by clipping the scene with a pre-defined scene boundary.

\begin{figure*}[t!]
    \centering
    \includegraphics[width=0.9\linewidth]{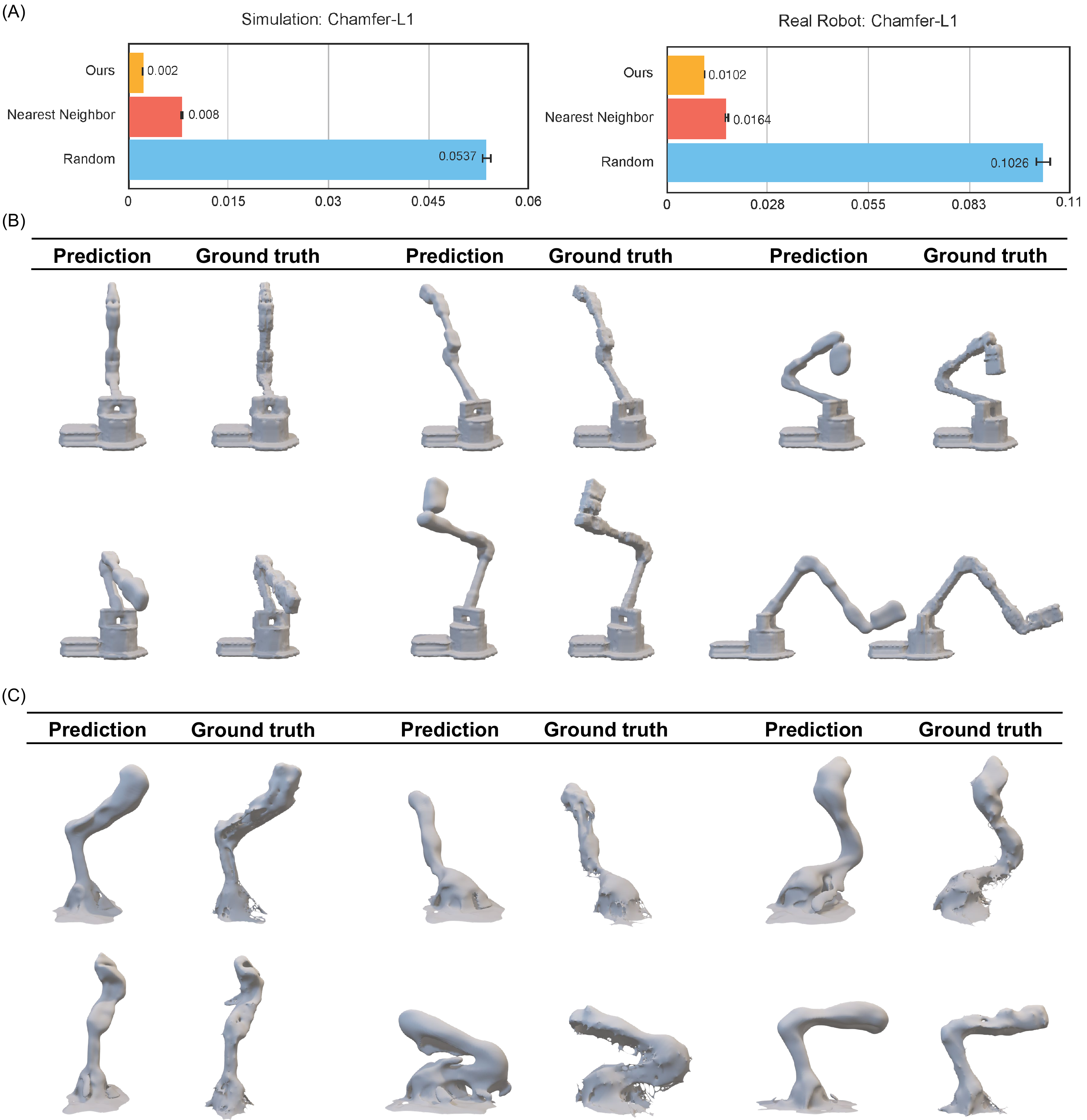}
    \caption{\textbf{visual self-model predictions.} (A) Quantitative evaluations of our visual self-model predictions in both simulated and noisy real-world environments. Our visual self-model outperforms nearest neighbor and random baselines suggesting that the visual self-model learns a generalizable representation of the robot morphology beyond the training samples. (B) With simulated training data, our visual self-model can produce high quality 3D body predictions given a diverse set of novel joint angles. (C) When the training data becomes highly noisy in the real world due to imprecise depth information, noisy camera calibrations and super sparse view points, our visual self-model can still accurately match the ground truth to reflect the overall robot body morphology and kinematics. See S3 Movie and S4 Movie for more examples.}
    \label{fig:prediction-results}
\end{figure*}

During data collection, we randomly moved the robot arms to get pairs of joint values and its corresponding point cloud. For each pair of data, the simulation needed less than 1 second and the real-world collection took around 8 seconds. In total, we collected 10,000 data points in simulation with PyBullet \cite{coumans2016pybullet} and 7,888 data points in the physical setup. We partitioned the data into training set ($90\%$), validation set ($5\%$) and testing set ($5\%$).

To evaluate the prediction accuracy, we ran several forward passes on the learned visual self-model to obtain the whole body mesh of the robot on the testing set. On a single GPU (NVIDIA RTX 2080Ti), this process took about 2.4 seconds. Following previous works on implicit neural representations of 3D models \cite{park2019deepsdf, sitzmann2020implicit}, we calculated the Chamfer-L1 distance between the predicted mesh and the ground truth mesh as our metric. All units in our paper are in meter.

In simulation, the point cloud fusion was nearly perfect due to noiseless depth image and exact camera calibrations. In the real-world experiments, we noticed that the point cloud fusion was very noisy due to imprecise depth information introduced by internal sensor errors, noisy camera calibrations, and importantly, very sparse view points. We did not increase the number of views since the current ground truth scan can already reflect the overall pose of the robot, so we tested the fidelity of our algorithm directly on the noisy real-world data in exchange of adding more resources and time cost. The gap of the ground truth data quality between the simulation and real world suggests that the final results in the real-world setup can be greatly improved with better future 3D scanning techniques.

Fig.\ref{fig:prediction-results} (B) visualizes pairs of predicted meshes and the ground truth meshes. In both simulation and real-world cases, our learned visual self-model produced accurate estimations of the robot morphology and kinematics, given only unseen joint angles as input. We also compared our algorithm with a random search baseline and a nearest neighbor baseline. For the random search, we randomly selected a robot mesh from the training set as the prediction. For the nearest neighbor baseline, we compared the testing joint angles with all the joint angles in the training set using L2 distance metric, and then used the robot mesh corresponding to the closest joint angles as the final prediction.

We presented the quantitative results in Fig.\ref{fig:prediction-results} (A). Our method outperforms both baselines suggesting that the our visual self-model learns the generalizable correspondence between the joint angles and the robot morphology as well as kinematics rather than memorizing the training set distribution.

\begin{figure*}[t!]
    \centering
    \includegraphics[width=\linewidth]{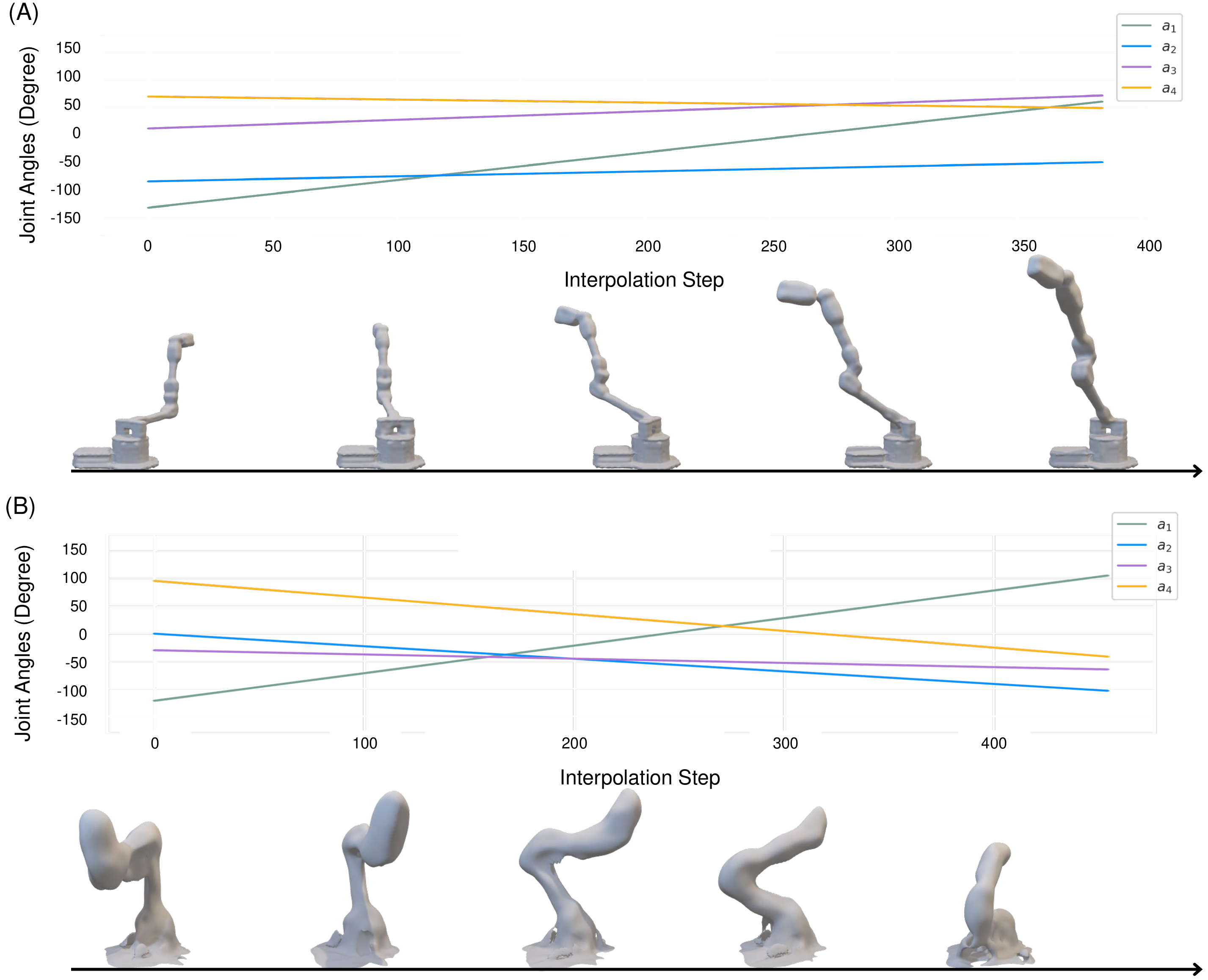}
    \caption{\textbf{Interpolation between joint angles.} We demonstrate that our learned visual self-model can smoothly interpolate between different joint angles. (A) shows the results trained in simulation and (B) shows the results trained on real-world data. See S5 Movie for more examples.}
    \label{fig:interpolation-results}
\end{figure*}

In addition to the predictions on individual set of joint angles, we also visualize the predictions over joint angle trajectories by linearly interpolating between sets of starting joint angles and sets of target joint angles. Both the starting and target joint angles are randomly sampled. As shown in Fig.\ref{fig:interpolation-results}, our visual self-model can generate smooth interpolations of robot morphologies between small changes of joint angles. As we will show next, this property allows our visual self-model to generate accurate trajectories for downstream motion planning tasks.

\subsubsection*{3D Self-Aware Motion Planning}

\begin{figure*}[t!]
    \centering
    \includegraphics[width=0.95\linewidth]{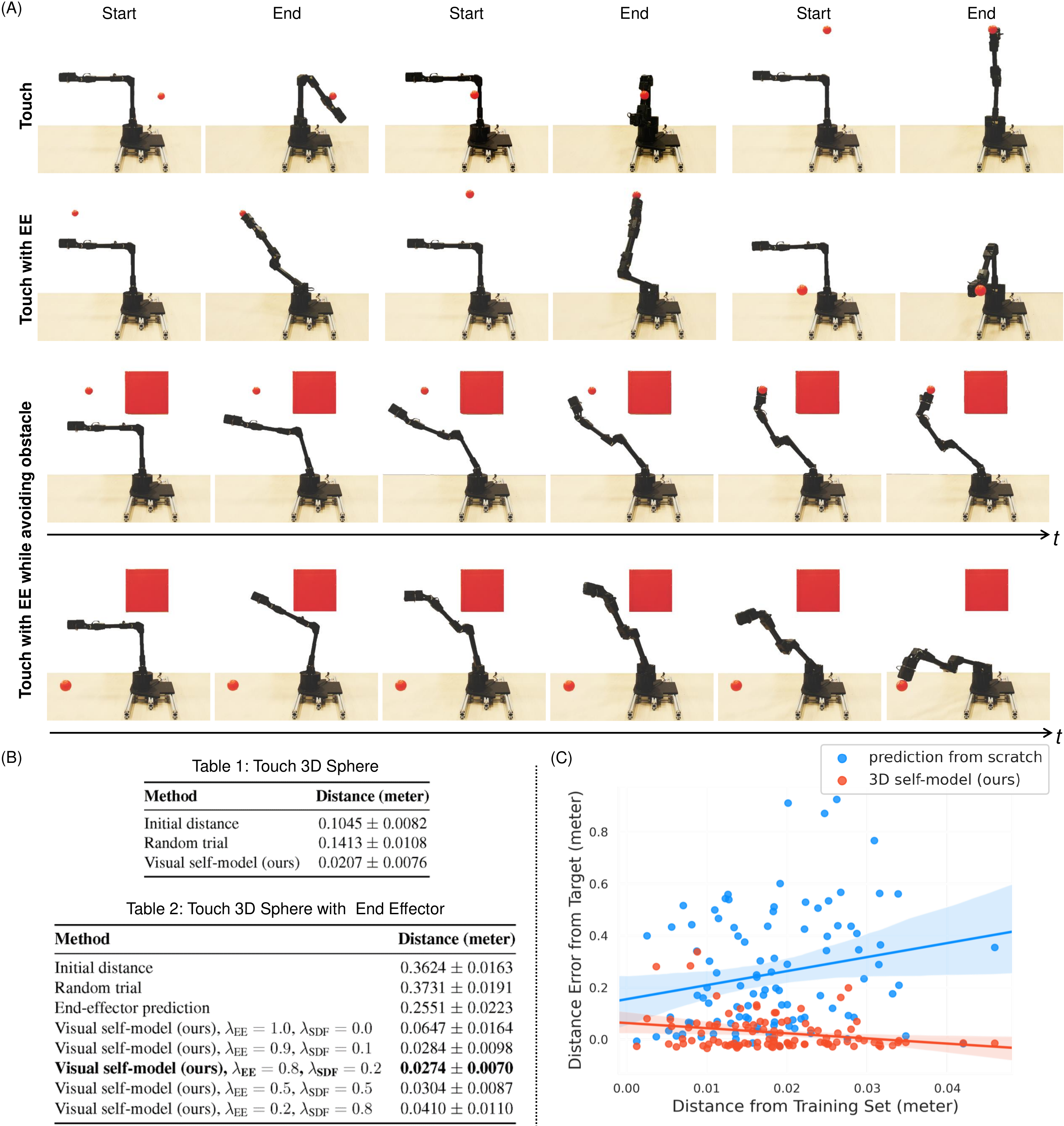}
    \caption{\textbf{3D self-aware motion planning results.} (A) For each of the three tasks, we show the real-world demos by executing the proposed plans from our visual self-model. See S6 Movie for more examples. (B) Our visual self-model outperformed all the baselines by a large margin. Overall, our visual self-model can produce accurate solutions for both tasks. (C) We found that our visual self-model enables the kinematic network to gain better generalization performance on downstream tasks than a plain kinematic self-model trained from scratch.}
    \label{fig:motion-planning-results}
\end{figure*}

In this subsection, we aim to evaluate the performance of using our visual self-model and 3D Self-Aware motion planning algorithms for three representative downstream tasks: teach a 3D sphere with any part of the robot body, touch a 3D sphere with end effector and touch a 3D sphere with end effector while avoiding obstacle. Detailed illustrations of the tasks and algorithms have been discussed above. For all three tasks, we present qualitative visualizations of our solutions obtained through the visual self-model in the real-world system in Fig.\ref{fig:motion-planning-results}. We then introduce our quantitative evaluation results in the simulation setup.

For the ``Touch a 3D sphere with any part of the body'' task, our evaluation metric measures the Euclidean distance of the closest points between the robot surface and the target object surface. We sampled 100 tasks where the target sphere is placed at different 3D locations within the reachable space of the robot. If the robot was already in contact with the target sphere at initialization, we discarded that task and re-sampled another task. Our results are shown in Fig.\ref{fig:motion-planning-results}(B). We compared our visual self-model with several other baselines. To reflect the task difficulty, we first measured the initial distance between the robot surface body at its home location and all sampled target sphere surfaces. We also compared with a random trial baseline where the only input was also the joint angles, similarly to our visual self-model. In this case, the robot randomly selected a set of joint angles as its final solution. This baseline gave even worse performance than initial distance indicating that the robot needs to perform careful inverse kinematics calculation with considerations of its entire morphology and kinematics. Overall, our method produces much more accurate solutions. Furthermore, our method was also time efficient during the search stage. Each solution took 2.92 seconds on average on a single GPU after 500 optimization iterations.

For the ``Touch a 3D sphere with end effector task'', our evaluation metric measures the Euclidean distance between the end effector link and the closest point on the target sphere surface. We sampled 100 tasks and made sure that the robot was not in contact with the target sphere at its home configuration. Our results are shown in Fig.\ref{fig:motion-planning-results}(B). Similar to the first task, we compared our approach with the initial distance and random trail baselines. Both baselines were poor at this task with about 36 cm to 37 cm errors. This is even worse than the first task because the presented task requires more accurate solutions to consider both the 3D body geometry and the end effector position.

We have hypothesized that our visual self-model encourages strong semantic knowledge of robot kinematics in the kinematic sub-network. To verify this hypothesize, we re-used the pre-trained weights of the kinematic sub-network, and appended two nonlinear layers of MLPs to perform further training only on the newly added layers, in order to regress the end effector link position. The quantity of the data and the strategy of data splits followed the same definition with our original visual self-model. The test error was around $0.5$cm. We also trained a network with the exact same architecture without pre-trained weights from our visual self-model to predict the end effector position. The test error of this model was $1.3$cm which was nearly three times higher. Moreover, when applying these two models separately with our motion planning pipeline in Fig.\ref{fig:motion-planning-results}(B), our method reached nearly ten times higher accuracy than the model trained from scratch denoted as ``end-effector prediction'' in the table. These results suggest the importance of considering the kinematic structure of the robot together with its 3D morphology. In terms of time efficiency, our method took $4.93$ seconds on average on a single GPU after 500 optimization iterations because of the fast parallel inference property.

Furthermore, we found that learning the kinematic structure, together with our visual self-model to learn the entire robot morphology, brought stronger generalization capability to downstream tasks. In Fig.\ref{fig:motion-planning-results}(C), every dot represents a task sample. The y-axis indicates the error measurements on the task of ``touch a 3D sphere with end effector'', and the x-axis denotes the closest distance between each sampled task and their nearest neighbor in the training set. Larger values on the x-axis means that the sampled task is farther away from the training data distribution. Therefore, the errors of the method with strong generalization capability should not raise with the increased distance from the training data distribution. We thus also plotted a linear regression model fit in the same figure. By comparing our visual self-model denoted as red dots and the model trained from scratch indicated as blue dots, we can tell that our visual self-model obtains a much stronger generalization capability, while the model trained from scratch will have a much higher error when the data is away from the training set.

Finally, we also provide results of using different values of $\lambda_{\textrm{Link}}$ and $\lambda_{\textrm{SDF}}$ in the objective function. We found that $\lambda_{\textrm{Link}}=0.8$ and $\lambda_{\textrm{SDF}}=0.2$ gives the best results. Therefore, adding a small regularization with the original SDF objective can help achieve better performance in this task.

For the ``Touch a 3D sphere with end effector while avoiding obstacle'' task, since the target joint states are generated and evaluated through the above task, we are now interested in evaluating the capability of generating collision-free trajectory when combing existing motion planners with our visual self-model as collision prediction function. Again, we sampled 100 tasks with initial states being contact free with the robot body. We placed a block $40$cm above the robot base as the obstacle object. The block has a dimension of $20$cm $\times$ $20$cm $\times$ $20$cm. In total, after running the motion planner with our visual self-model, we received 95 out of 100 trajectories which the model believes no collision will happen along each trajectory. We then executed these trajectories and found that 92 out of the 95 trajectories successfully passed around the obstacle towards the target object without any collision. This is $96.84\%$ success rate over all the output trajectories. Our method took $7.43$ seconds on average to produce an entire trajectory which includes the time for both inferring the target state as presented in the second task and running the motion planners. This fast inference time enables our method to provide real-time planning and control solutions.

\subsubsection*{Resiliency Tests}

\begin{figure*}[t!]
    \centering
    \includegraphics[width=0.75\linewidth]{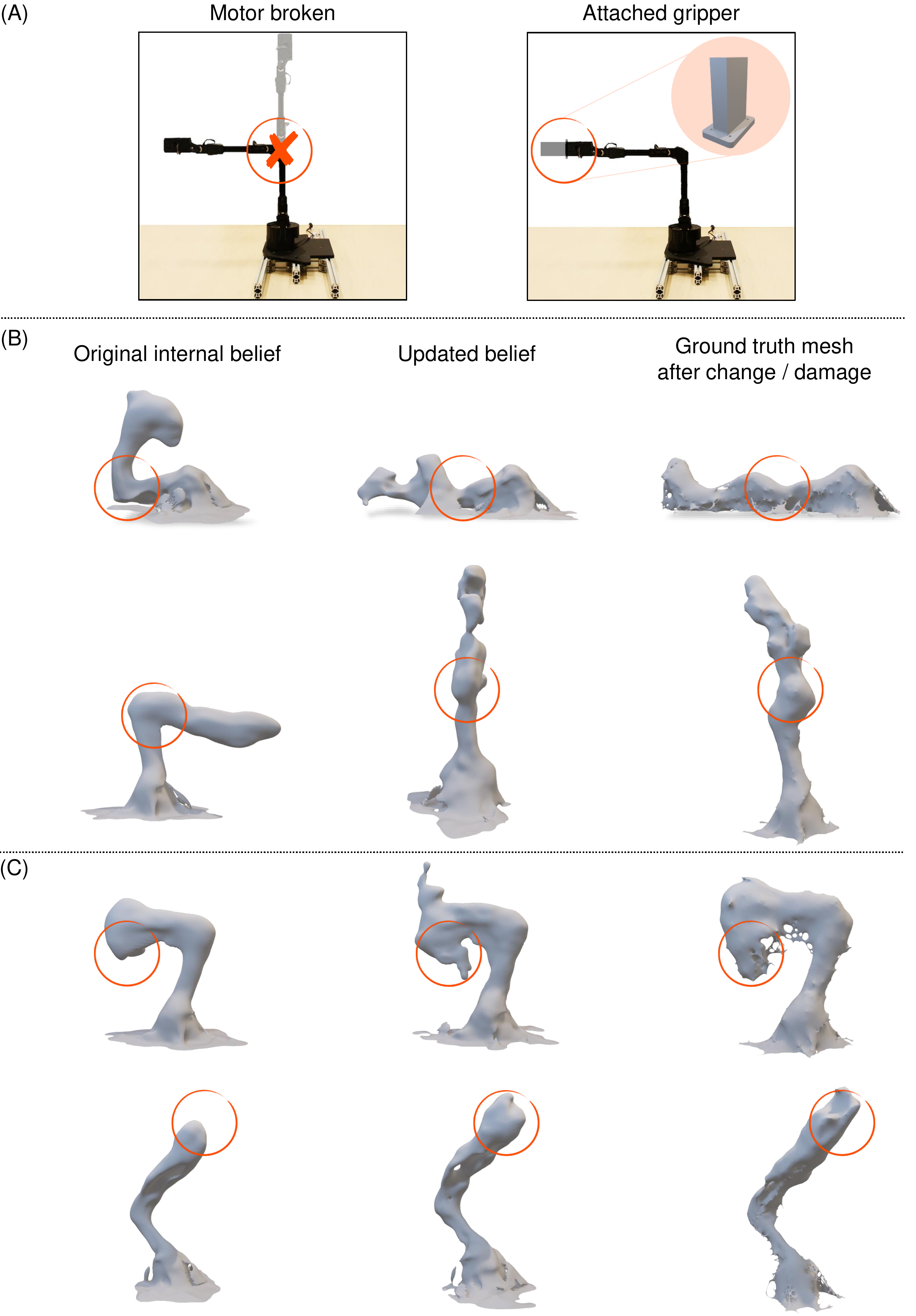}
    \caption{\textbf{Potential change or damage on the robot and visualizations} (A) Two types of potential changes. The left scenario is motor broken where the joint will always stay at $90\degree$. In the right scenario, we attached a 3D-printed plastic stick. (B) Motor broken: we can visualize the robot's original internal belief, its updated belief after continual learning and the current robot morphology. (C) Extended robot link visualizations.}
    \label{fig:damage-illustration}
\end{figure*}

Being able to identify potential damages or changes to the robot body and quickly recover from these changes is a critical capability of intelligent machines in the real world. We made two type of changes to the robot body as depicted by Fig.\ref{fig:damage-illustration}(A). In the first change, we broke the second motor to the end effector link by disconnecting the data transfer cable from the motor, which results the corresponding joint always stayed at $90\degree$. Motor broken can happen due to various reasons such as loosing cables, over heating or hardware damage, but the common resulted observation is that the motor does not respond to any commands. The second change applies to the topology change of the end effector link. We attached a 3D-printed plastic stick to the end effector so that the reachable space of the robot arm was extended. This is also a representative change in practical applications when different tasks demand new attachments of tools to the robot body or switch different grippers on a robot arm.

\begin{figure*}[t!]
    \centering
    \includegraphics[width=0.9\linewidth]{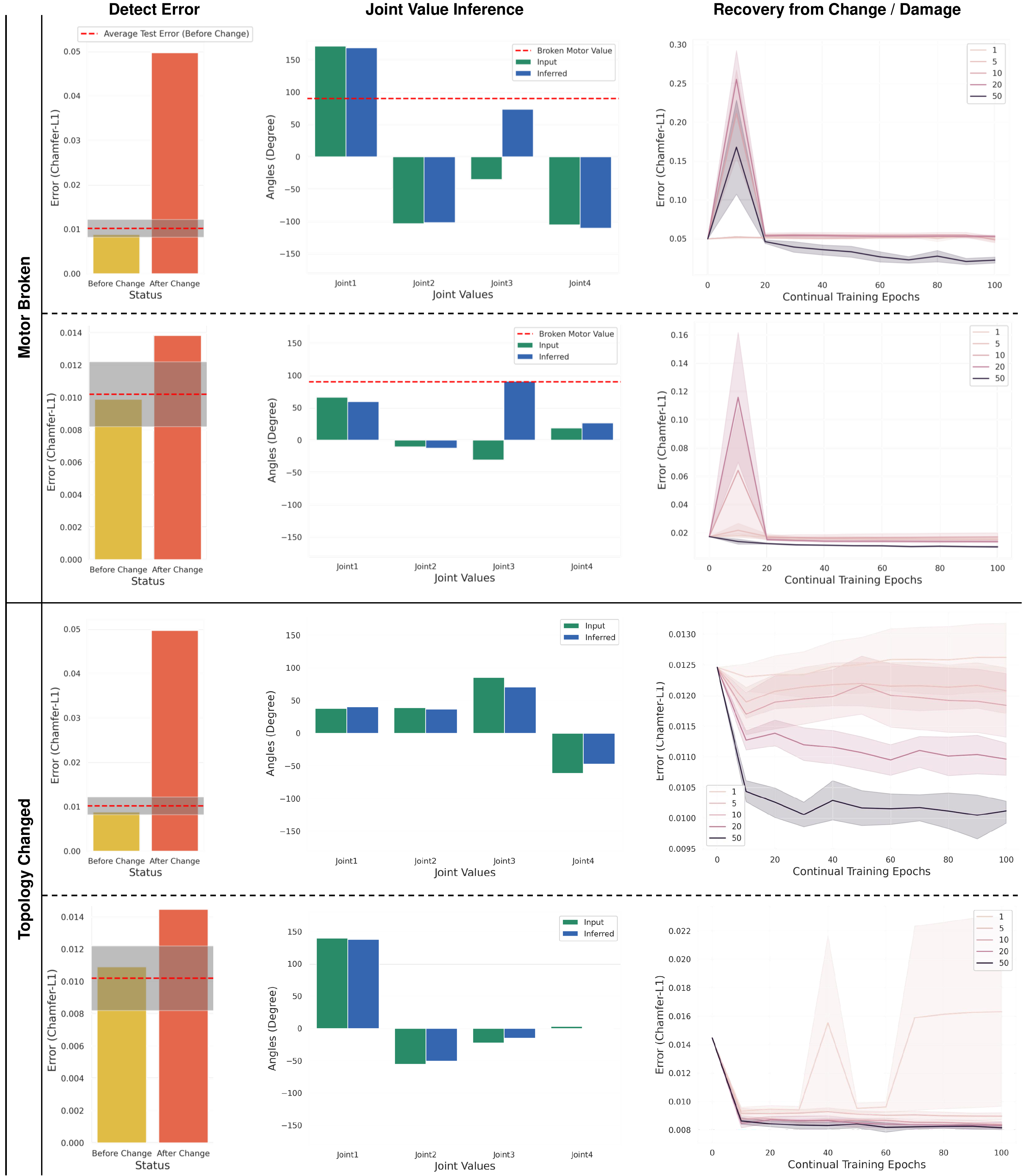}
    \caption{\textbf{Resiliency Tests.} In the first column, the learned visual self-model can detect the change or damage through the large error gap. In the middle column, the learn3d visual self-model can identify the specific type of change through the mismatch between the input joint values and the inferred joint values. In the last column, we show how the visual self-model can update its internal belief to match the current robot morphology.}
    \label{fig:resiliency-test}
\end{figure*}

With our proposed algorithm and the learned visual self-model, we tested the applicability of our method directly on these real world changes. Fig.\ref{fig:resiliency-test} presents several example results. The first step is to detect the change. As shown in the first column, our algorithm detected a clear gap between the original prediction errors and the current prediction errors. The obvious gaps suggest that our visual self-model can capture the changes happening on the robot body.

The second step is to identify the specific type of change. In the first two examples, no matter what the input commands were to the robot, the second last joint was always inferred to be around $90\degree$ by solving the inverse problem with the newly observed morphology. This consistent mismatch indicates that the second last motor was broken and the angle stayed at $90\degree$. In the last two examples, even though we can detect that there were some changes from the first step, the inferred joint angles were still well-aligned with the input commands. Following our discussions earlier, our algorithm identified that there was a topology change on the robot body. Our results suggest that our visual self-model can be used to effectively solve inverse problems to help identify what body change or damage might have taken place. Importantly, our approach only requires a single 3D observation of the current robot to produce the above results to detect and further identify the damage.

In the final step, our goal is to evaluate whether our visual self-model can quickly recover from the detected changes with only a few new observations. We first collected a few more observations of the current robot through random movements. With the new observations, we used them as the training data to continue the training of our existing visual self-model. Fig.\ref{fig:resiliency-test} plots the intermediate model performances on the test instance at every 10 epochs. We found that our model required $50$ examples to converge. Our visual self-model can quickly recover with the new training data after 100 epochs which took on average $8.13$ minutes in the real world on a single GPU.

Another advantage of our visual self-model is its interpretability. In Fig.\ref{fig:damage-illustration}(B), we can visualize the internal belief of the robot before and after the damage adaption. Through these visualizations, we can inspect what the robot's internal belief looks like and whether the robot has successfully updated its belief to match the current robot morphology. These visualizations can be queried in an online fashion with about $2.4$ seconds on a single GPU.

\subsection*{Discussion}

We have introduced Neural 3D Visual Self-Model and algorithm designs on how to leverage it for 3D motion planning and control tasks. We have also constructed a pipeline to demonstrate the resiliency of the visual self-model from damage detection, identification and recovery. These innovations have equipped our visual self-model with several special properties that are particularly useful in real-world robotic applications.

It is important to model the robot morphology in continuous 3D domain. We realized this with the implicit representations of the 3D shape where the input spatial coordinates and joint angles are both continuous. Thus, at inference time, one can query the 3D information of the robot morphology and kinematics at arbitrary spatial location given any joint angles in a highly memory efficient manner where the only storage cost is the weights of the network which is 1.1M. The queried resolution can also vary depending on the precision required for different tasks. Since the entire 3D robot morphology is modelled, the task solutions provided by our visual self-model can always consider relevant body part geometries when different parts of the robot interact with the environment.

In order to better model the kinematic structure of the robot, learning the kinematic features of the robot together with its 3D morphology can be very helpful. Due to the decoupling of the spatial information distillation and kinematic information distillation, we can obtain a kinematic branch that explicitly learns the robot's kinematic structure. As we have shown in our ablation study, the kinematic branch captured precise end effector position given input joint angles. In fact, the kinematic branch trained together with the final SDF prediction produced more accurate predictions than a specialized network trained from scratch for end effector predictions. This suggests the importance of explicitly considering the entire robot morphology.

Furthermore, making the entire visual self-model differentiable can speed up the planning process. Not only our entire visual self-model is differentiable, the model can also be queried at body parts instead of the entire body. The differentiability of the model allows us to easily perform back-propagation with respect to the input joint angles to solve inverse problems. Since the model can be queried with only subset of inputs that are of our current interests depending on the task, we only need to spend computational resources and time on the task-relevant components. Both benefits make the model super easy and efficient to work with. Lastly, our model can be easily distributed on GPUs. With one single GPU, our experiments already achieved highly efficient planning. With more computing resources, we expect that our model can reach even faster inference speed.

There are several opportunities to improve our current approach in future works. Though our visual self-model runs fast for downstream task planning, the training costs about a day to obtain high quality results on a single GPU. For applications requiring faster convergence, the training time could potentially be reduced by applying meta-learning \cite{tancik2021learned} techniques to get better initialization of the neural network weights with a subset of the training data. Another possible solution is to employ an exploration policy \cite{goldberg1984active, bongard2005automatic, bohg2017interactive, ramakrishnan2019emergence} to select informative data samples over uniform random sampling, which may lower the total number of training data needed to obtain faster training speed and higher data efficiency.

The second improvement can be noticed from the precision gap between the ground truth data from the simulation environment and the ground truth data from the real-world scans. As we have discussed in the Results section, this gap is caused by imprecise depth data, noisy camera calibrations and super sparse camera views. The current real data quality may not handle very fine-grained details. It is possible to improve the data quality with more dedicated depth sensors or more camera views. However, the dedicated 3D scanners often cost ten to hundreds of times more than our current solution. In our problem formulation, they also suffer from much slower scanning speed, limited scanning range and human efforts to manually posit the scanners around the robot. Another potential solution is to use more camera views. Therefore one may think of using structure from motion and multi-view stereo framework \cite{schoenberger2016sfm, schoenberger2016mvs} to reconstruct the 3D model of the robot. While these state-of-the-art techniques can provide high quality 3D reconstructions, in our trail on a GPU workstation, they required dozens or even hundreds of camera views and hours of processing time to obtain a single scan, which makes it difficult to scale these approaches to our problem setup. As a comparison, our current pipeline takes a few seconds to obtain a complete 3D mesh with only five camera views. The ideal solution should be both fast and accurate without too much human supervision during the scanning phase. Future progresses on hardware and software improvements along this direction can substantially improve the real-world data quality.

Overall, our method opens up a new opportunity to learn a visual self-model of robots that is 3D-aware, continuous, memory efficient, differentiable and kinematics-aware for fast motion planning and control, with potentials to scale to other robotic platforms and applications such as locomotion and object interaction.

\subsection*{Materials and Methods}

Our visual self-model is consisted of three neural network components: a coordinate network, a kinematic network, and a network to fuse the coordinate features and kinematic features to produce the final SDF. The coordinate network is a single layer of MLP, and the kinematic network has four layers of MLPs. The output features from these two networks are concatenated along the feature dimension. The concatenated features are then sent into another four layers of MLPs to output the final SDF value. We used sine function as non-linear activations throughout the entire network to obtain high resolution details and initialized the network weights to preserve the distributions of the activations. We optimized the network for 2,000 epochs with Adam \cite{kingma2014adam} optimizer, and we implemented the entire network with PyTorch \cite{NEURIPS2019_9015} and PyTorch Lightning \cite{Falcon_PyTorch_Lightning_2019} framework. Our training used the batch size of $1.536 \times 10^5$ and the learning rate of $5 \times 10^{-5}$ on a single NVIDIA RTX 2080 Ti GPU. All the input data were normalized to have zero mean and a range of $[-1, 1]$.

We followed previous work to minimize the following loss function when predicting the SDF value as an Eikonal boundary value problem \cite{sitzmann2020implicit}:
\[L_{SDF} = \int_{\Omega} \vert\vert\ |\nabla_{\vb*{I}} H(\vb*{I}) | - 1 \vert\vert \mathop{d\vb*{I}} + \int_{\Omega_0} \vert\vert H(\vb*{I}) \vert\vert + (1 - \langle \nabla_{\vb*{I}} H(\vb*{I}), \textbf{n}(\vb*{X}) \rangle) \mathop{d\vb*{I}} + \int_{\Omega \setminus \Omega_0} \psi(H(\vb*{I})) \mathop{d\vb*{I}}, \] where $\vb*{I} = (\vb*{X}, \vb*{A})$ is the concatenation of the input coordinates and joint angles, $H = F \circ (C, K)$ and $\psi(\vb*{I}) = \textrm{exp}(-\alpha \cdot \vert H(\vb*{I}) \vert)$. $\Omega$ represents the whole spatial domain and $\Omega_0$ denotes the zero-level set. In total, there are three terms that sums up together to get the final loss. The first term constraints the norm of the spatial gradients of the on-surface points to be one. The second and the third term separately encourages the on-surface points and off-surface points to follow the definition of zero-level SDF. The on-surface points should stay close to zero values and ground truth normals, while the off-surface points should not have close to zero SDFs. During training, we sampled the same number of points for both on-surface and off-surface scenarios for every batch.

\textbf{Funding:} This research was supported by DARPA MTO Lifelong Learning Machines (L2M) Program HR0011-18-2-0020, NSF NRI Award \#1925157, NSF CAREER Award \#2046910, and a gift from Facebook Research.\textbf{Author contributions:} B.C. and H.L. proposed the research; B.C. developed the main idea, algorithm designs, implementations, simulation and hardware experiments. H.L. and C.V. provided deep insights and guidance on the algorithm and experiment design. B.C., H.L., and C.V. performed numerical analysis. R.K. provided help on hardware experiments and was involved in the discussions. B.C., H.L., and C.V. wrote the paper; all authors provided feedback. \textbf{Data and materials availability:} We will open source all data and software codebase needed to evaluate the conclusion in the paper, the Supplementary Materials and our project website. Additional information can be addressed to B.C.

\bibliography{scibib}

\bibliographystyle{Science}

\clearpage

\end{document}



\baselineskip24pt


\maketitle


\begin{table}[]
\centering
\begin{tabular}{@{}lc@{}}
\toprule
\textbf{Method}      & \textbf{Distance (meter)} \\ \midrule
Initial distance     & 0.1045 $\pm$ 0.0082       \\
Random trial         & 0.1413 $\pm$ 0.0108       \\
Visual self-model (ours) & 0.0207 $\pm$ 0.0076       \\ \bottomrule
\end{tabular}
\caption{reach3d}
\label{tab:reach-3d}
\end{table}

\begin{table}[]
\centering
\begin{tabular}{@{}lc@{}}
\toprule
\textbf{Method}                                                                          & \textbf{Distance (meter)}    \\ \midrule
Initial distance                                                                         & 0.3624 $\pm$ 0.0163          \\
Random trial                                                                             & 0.3731 $\pm$ 0.0191          \\
End-effector prediction                                                                  & 0.2551 $\pm$ 0.0223          \\
Visual self-model (ours), $\lambda_{\textrm{EE}}=1.0$, $\lambda_{\textrm{SDF}}=0.0$          & 0.0647 $\pm$ 0.0164          \\
Visual self-model (ours), $\lambda_{\textrm{EE}}=0.9$, $\lambda_{\textrm{SDF}}=0.1$          & 0.0284 $\pm$ 0.0098          \\
\textbf{Visual self-model (ours), $\lambda_{\textrm{EE}}=0.8$, $\lambda_{\textrm{SDF}}=0.2$} & \textbf{0.0274 $\pm$ 0.0070} \\
Visual self-model (ours), $\lambda_{\textrm{EE}}=0.5$, $\lambda_{\textrm{SDF}}=0.5$          & 0.0304 $\pm$ 0.0087          \\
Visual self-model (ours), $\lambda_{\textrm{EE}}=0.2$, $\lambda_{\textrm{SDF}}=0.8$          & 0.0410 $\pm$ 0.0110          \\ \bottomrule
\end{tabular}
\caption{reach3d\_link}
\label{tab:reach-3d-link}
\end{table}

\clearpage


\bibliography{scibib}

\bibliographystyle{Science}

\clearpage